\documentclass[conference]{IEEEtran}
\IEEEoverridecommandlockouts
\usepackage{cite}
\usepackage{amsmath,amssymb,amsfonts}
\usepackage{algorithmic}
\usepackage{graphicx}
\usepackage{textcomp}
\usepackage{xcolor}
\usepackage{nomencl}
\usepackage{makecell}
\usepackage[normalem]{ulem}
\usepackage{hyperref}

\renewcommand\arraystretch{1.2}

\def\BibTeX{{\rm B\kern-.05em{\sc i\kern-.025em b}\kern-.08em
    T\kern-.1667em\lower.7ex\hbox{E}\kern-.125emX}}
\begin{document}

\title{Range~and~Angle~Estimation with Spiking~Neural~Resonators for FMCW~Radar\\
%\thanks{}
}

\author{
\IEEEauthorblockN{Nico Reeb\IEEEauthorrefmark{1}, %\orcidlink{0000-0003-4617-6215}, 
 Javier Lopez-Randulfe, %\orcidlink{0000-0002-6922-6951},
 Robin Dietrich %\orcidlink{0000-0001-5820-1399} 
 and Alois C. Knoll\\%\orcidlink{0000-0003-4840-076X}\\
\IEEEauthorblockA{\textit{TUM School of Computation, Information and Technology}}
\textit{Technical University of Munich}\\
Munich, Germany \\
\IEEEauthorrefmark{1}\href{mailto:nico.reeb@tum.de}{nico.reeb@tum.de}}
}

\maketitle
\pagestyle{plain}

\begin{abstract}

Automotive radar systems face the challenge of managing high sampling rates and large data bandwidth while complying with stringent real-time and energy efficiency requirements. 
The growing complexity of autonomous vehicles further intensifies these requirements.
Neuromorphic computing offers promising solutions because of its inherent energy efficiency and parallel processing capacity.

This research presents a novel spiking neuron model for signal processing of frequency-modulated continuous wave~(FMCW) radars that outperforms the state-of-the-art spectrum analysis algorithms in latency and data bandwidth.
These spiking neural resonators are based on the resonate-and-fire neuron model and optimized to dynamically process raw radar data while simultaneously emitting an output in the form of spikes. 
We designed the first neuromorphic neural network consisting of these spiking neural resonators that estimates range and angle from FMCW radar data.
We evaluated the range-angle maps on simulated datasets covering multiple scenarios and compared the results with a state-of-the-art pipeline for radar processing.

The proposed neuron model significantly reduces the processing latency compared to traditional frequency analysis algorithms, such as the Fourier transformation~(FT), which needs to sample and store entire data frames before processing.
The evaluations demonstrate that these spiking neural resonators achieve state-of-the-art detection accuracy while emitting spikes simultaneously to processing and transmitting only 0.02\% of the data compared to a float-32 FT. 
The results showcase the potential for neuromorphic signal processing for FMCW radar systems and pave the way for designing neuromorphic radar sensors.
\end{abstract}

\fbox{\begin{minipage}{0.9\columnwidth}
This is the version of the article before peer review or editing, as submitted by an author to \textit{Neuromorphic Computing and
Engineering}. IOP Publishing Ltd is not responsible for any errors or omissions in this version of the manuscript or any version derived from it. 
%The Version of Record is available online at [INSERT DOI]
\end{minipage}}
\section{Introduction}

As society and industry depend on increasingly complex signal processing systems, solutions become more energy-intensive.
This trend drives the need for optimized processing pipelines that minimize energy consumption and
data bandwidth while improving performance and reliability \cite{leiserson2020there}.
The automotive industry is paradigmatic due to the limited energy availability in cars and the need for precise and reliable low-latency systems to achieve fully autonomous driving \cite{lin2018architectural, vogginger2022}.
One of the most critical operations in advanced driving assistance systems (ADASs) is sensor signal processing. Subsequent crucial tasks, such as navigation, collision avoidance, or adaptive cruise control, depend on accurate and fast sensor data processing.
The frequency-modulated continuous-wave radar (FMCW) sensor is a prominent example utilized in ADASs.
Its low cost and long-range detection, combined with its robustness to bad lighting and weather conditions, make it a fundamental element for perceiving the car's surroundings \cite{patole2017automotive}.

A promising approach to developing low-power processing systems is to mimic biology.
The human brain can collect and process sensor data, reason, and decide how to interact with the environment while consuming only 20 W \cite{sokoloff_1996}.
The brain's superior efficiency arises from parallel, asynchronous, and event-driven processing, and communication via binary spikes.
Neuromorphic computing is a research field that aims to replicate these characteristics of the brain to create accurate and energy efficient solutions to real-world problems \cite{mead2020we,markovic2020physics}.
Spiking neural networks (SNNs) are among the most promising neuromorphic algorithms \cite{mass1997}.
Neuromorphic hardware is developed in symbiosis to take advantage of the efficient processing of neuromorphic algorithms.
SNNs consist of an asynchronous network of artificial neurons that transmit information via spikes, relying on internal dynamics models driven by a continuous stream of incoming data.
In addition to advancing the understanding of the human brain \cite{rhodes2020real}, research in SNNs focuses on engineering problems, such as finding an efficient replacement for deep neural networks (DNNs) \cite{rueckauer2017conversion}, optimizing operations in large data centers \cite{vogginger2024neuromorphic}, or designing algorithms for small embedded systems \cite{bing2018survey}.
Most neuron models in SNNs follow integrate-and-fire dynamics, where a neuron charges over time based on input activity and emits a spike once its state crosses a defined threshold.
IF models are present in most neuromorphic chips \cite{davies2018loihi, rhodes2018spynnaker}, and offer good results in the conversion of DNNs \cite{rueckauer2017conversion}, applying biologically inspired learning techniques \cite{kheradpisheh2018stdp}, or solving complex optimization problems\cite{davies2018loihi}.

Object detection with FMCW radars is based on the frequency analysis of the continuous analog signal generated by the sensor.
Traditional signal processing methods typically convert discrete time sampled data into the frequency domain, with the Fourier transform (FT) being the most widely used technique for frequency analysis.
Some neuromorphic algorithms have already emerged for computing the frequency spectrum.
Authors in \cite{jimenez2016binaural} apply sequential spiking band-pass filters to process audio signals based on their frequency spectrum.
The works in \cite{lopez2022time} and \cite{lopez2023integrate} introduced an electric circuit and a neuron model that encode analog signals to temporal spikes and provide a mathematically equivalent representation of the FT.
Alternatively, Izhikevich proposed the resonate-and-fire (RF) neuron model, which enables frequency analysis on continuous input data \cite{izhikevich2001:rfneurons}. 
The researchers in \cite{orchard2021, auge2020selective, hille2022} studied the application of resonate-and-fire neurons to oscillating signals.
Approaches utilizing resonate-and-fire neurons focused on one-dimensional frequency analysis, whereas FMCW radar processing relies on multi-dimensional frequency analysis for angle and velocity information.
To the best of our knowledge there is no research on how resonating neurons compare with classic frequency analysis methods, such as the FT, on multi-dimensional FMCW radar data.

In this work, we advance the resonate-and-fire neuron model \cite{izhikevich2001:rfneurons} to concurrently process incoming radar data and optimally convey information through spiking. 
Traditional frequency analysis techniques, such as the FT, calculate the spectrum after storing all data samples in a given time window.
By allowing concurrent processing and spiking, we can achieve three benefits over traditional approaches:
\begin{itemize}
    \item[1)] Reduce the latency of target detection, as each data sample incoming updates the estimate.
    \item[2)] Reduce data bandwidth due to sparse spiking.
    \item[3)] Remove the need for memory storage for sensor data due to the immediate processing by the neurons.
\end{itemize}
We arranged these neural resonators into a single layer capable of performing two-dimensional frequency analysis on FMCW radar data allowing to estimate range and angle simultaneously (Section \ref{sec:angle_estimation}).
To maintain the accuracy achieved by state-of-the-art methodologies, we developed neuron dynamics that filter out noise from the incoming signal (\ref{sec:gradient}). We implemented three different spike functions including time- and rate-coding for comparison (Section \ref{sec:spiking_functions}).
We simulated radar datasets with exact positional information of point targets, a level of accuracy not available in real radar datasets, 
validated the approach on these datasets, and compared it with the FT method (Section \ref{sec:evaluation}). 
Finally, we applied the spiking neural resonators to public radar datasets, allowing us to visually assess the model's generalization capabilities (Section \ref{sec:real_data}).
We implemented the neuron model on GPU and published the code \footnote{\url{https://github.com/ndotr/Spiking-Neural-Resonator-Network}}.
\section{Neuron model and network architecture}
\label{sec:model}

\begin{figure}[t]
    \centering
    \includegraphics[width=\columnwidth]{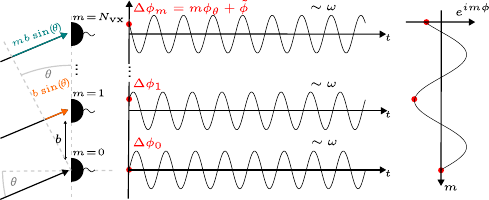}\vspace{0.5cm}
    \caption{
    Geometrical visualization of the radar signals. On the left, antenna layout with $N_\text{vx}$ virtual antennas in one line and a spacing $b$ between consecutive antennas.
    Transmitted and reflected signals are indicated with arrows, and the direction-of-arrival (DoA) is given as angle $\theta$.
    In the middle, schematic view of temporal dynamics of the IF signals $x_m(t)$ with frequency $\omega$, where $m$ is the antenna index. 
    Frequency analysis along the temporal dimension provides information on the range of an object. 
    On the right, schematic view of the complex value $\exp(i m \phi)$ over virtual antennas. 
    Frequency analysis along the antenna dimension provides information on the DoA.
    }
    \label{fig:radar_data_vx}
\end{figure}

Typically, automotive radar systems rely on multiple-input-multiple-output FMCW radar sensors with multiple transmitting as well as receiving antennas. 
This enables a high range and angular resolution as well as unambiguous velocity detection.
Various multiplexing methods \cite{hamidi2018, hamidi2021, zwanetski2013} enable simultaneous transmission from multiple antennas while maintaining orthogonality between the signals. 
The combination and mixing of multiple transmit and receive antennas leads to \mbox{$N_\text{vx} = N_\text{tx} \times N_\text{rx}$} virtual antennas. 
The $m$-th virtual antenna returns an intermediate-frequency (IF) signal~$x_m(t)$, leading to a continuous data vector~\mbox{$\vec{x}(t) \in \mathbb{C}^{N_\text{vx}}$} for the entire radar sensor (see Fig. \ref{fig:radar_data_vx}).
Throughout the paper, we assume sawtooth frequency modulation and a simplified antenna layout with one transmitting antenna and $N_\text{rx} = N_\text{vx}$ receiving antennas in one line parallel to the ground.
Reflections of $K$ objects lead to an IF signal that can be modeled for one chirp \mbox{$t \in [0, T_c]$} as
\begin{align}
    x_m(t) &= \sum^K_k a_k e^{i m \phi_k} e^{i \omega_k t}.
\end{align}
The frequency $\omega_k$ is directly proportional to the radial~range~$r_k$ between the object~$k$ and the sensor. 
The phase shift $m \phi_k$ depends on the angle between object~$k$ and the sensor, and the position of the $m$-th antenna within the antenna layout.
The amplitude $a_k$ depends on various factors, such as the radar cross-section of the object, the distance between the radar and object $d_k$, the dynamics of the sensor \cite{hau2020}, and properties of the medium. 
More details on the fundamentals on radar signal processing can be found in \cite{fmcw_radar_designs}.

A single spiking neural resonator receives and continuously processes the radar signal vector $\vec{x}(t)$ (see Fig. \ref{fig:radar_data_vx}).
We describe the different processing steps of the neuron model in the following.
First, a complex weight matrix multiplication extract the angle information of the sensor data.
Second, the resonator dynamics provide information about the distance. 
Third, neuron dynamics analyze the temporal behavior of the resonator to produce informative spikes. 
We compare three spiking functions, relying on rate-coded and time-coded approaches, respectively.

\subsection{Angle estimation - Dendritic vector multiplication}
\label{sec:angle_estimation}

The relative phase shifts between the antenna signals of a multi-array antenna system (see Fig. \ref{fig:radar_data_vx}, red) contain information on the direction-of-arrival (DoA) of the objects.
Due to a much larger range $r_k$ compared to the distance $b$ between consecutive antennas ($r_k \gg b$), we can assume that the incoming signals are aligned parallel on all antennas.
Due to the geometric setup of the antenna array, the range $r_k$ between object $k$ and receiving antenna changes slightly for each antenna (see Fig. \ref{fig:radar_data_vx}). 
The relative displacements of the range between the $0$-th and $m$-th antenna \mbox{$\Delta d_k = m b \sin(\theta_k)$} depends on the angle $\theta_k$ between target $k$ and sensor, and on the antenna distance $m b$.
This displacement results in a phase shift between the IF signals of two antennas.
By assuming the antenna distance to match multiples of half of the wavelength of the radar sensor \mbox{$b = \lambda / 2$}, the relative phase shift between antenna signals $x_0(t)$ and $x_m(t)$ can be stated as \mbox{$m \cdot \phi(\theta_k) = m \cdot \pi \sin(\theta_k)$}.

Frequency spectrum analysis, such as the discrete FT (DFT), estimates the angle information in the signal of a multi-array antenna system.
The DFT is a matrix multiplication of a complex matrix $W$ with the IF signal vector $\vec{x}(t)$.
The matrix $W$ consists of matrix elements \mbox{$W_{l m} = e^{-i m \phi_l}$}, where \mbox{$l \in [0, N_\text{vx}]$} indicates phase shifts depending on the resolution of the DFT.
Accordingly, we rewrite the DFT as
\begin{align}
    y_{l}(t) &= \sum^{N_\text{vx}}_{m} W_{lm} x_m(t)\\
    &= \sum^{N_\text{vx}}_{n} e^{-i m \phi_k} \sum^K_k a_k e^{i m \phi_k} e^{i \omega_k t}\\
    &= \sum^K_k a_k \underbrace{\sum_m e^{im(\phi_k - \phi_l)}}_{\beta_{kl}} e^{i \omega_k t} \\
    &= \sum^K_k a_k \cdot \beta_{kl} \cdot e^{i \omega_k t}.
    \label{eq:angle_detection}
\end{align}

The case of matching phase shifts for a single target ($l = k$) results in $\beta_{kl}=1$ and hence in the highest amplification of the signal $y_k = N_\text{vx} a_k e^{i \omega_k t}$, whereas non-matching phase shifts ($l \neq k$) result in complex factors $\beta_{kl}$ with \mbox{$|\beta_{kl}| < 1$}. 
After matrix multiplication, high amplifications indicate an object at a given angle $\theta_l$.

In our proposed network, a single neuron $l$ performs a vector multiplication \mbox{$y_l(t) = \vec{w}_l \cdot \vec{x}(t)$} of the weight vector \mbox{$\vec{w}_l = (W_{l0}, ..., W_{lm})$} and the radar signal vector $\vec{x}(t)$, therefore each neuron corresponds to an angle $\theta_l$.
The resulting signal $y_l(t)$ is further processed in the neuron.

\subsection{Distance estimation - Neural resonators}

\begin{figure}[t]
    \centering
    \includegraphics[width=\columnwidth]{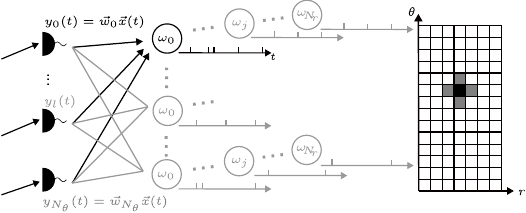}
    \caption{
    Network of spiking neural resonators. Reflected signals are indicated as arrows, detected by virtual antennas. 
    Each virtual antenna passes its IF signal $x_m(t)$ to a neural resonator with given eigenfrequency $\omega_j$ weighted by the complex weight vector $\vec{w}_l$.
    Along the vertical dimension, complex weight vectors optimized for a specific angle $\theta$ are visualized (gray, black).
    Along the horizontal dimension, the eigenfrequency of the neuron changes optimized for a specific range $r$.
    The transmitted spikes can be visualized as range-angle map.
    }
    \label{fig:architecture}
\end{figure}

A spectrum analysis along the temporal dimension extracts the range from a single chirp of the IF signal $x_m(t)$ or the weighted IF signal $y_l(t)$ of an FMCW radar as the frequency is directly proportional to the range of an object to the sensor.
We use the Resonate-and-Fire neuron model \cite{izhikevich2001:rfneurons, orchard2021} as oscillating neuron model,

\begin{align}
    \label{eq:de_neuron}
    \dot{s}_{jl}(t) = a e^{i \omega_j} s(t) + y_l(t),
\end{align}
with $j$ indexing the eigenfrequencies that represent different distances.
Typically, we set the initial value of $s$ at $s(0) = 0$ for each chirp. Assuming a general input \mbox{$y_l(t) = \sum_k^K a_k \beta_{kl} e^{i \omega_k t}$}, we can determine the analytic solution to the differential equation
(\ref{eq:de_neuron}),

\begin{align}
    \label{eq:sol_de_neuron}
    s_{jl}(t) = \sum_k^K \frac{i a_k \beta_{kl} e^{i w_j t}}{w_j - w_k} (e^{-i(w_j - w_k)t} - 1) .
\end{align}
To reduce computational costs, we use the discrete version of the neuron model in the following work 
and no decay by setting $a_{kl} = 1 \, \forall k$. 
Because an analog-to-digital converter (ADC) typically samples the radar data, the discrete version of the neuron model is sufficient.

\begin{align}
    \label{eq:discrete_neuron}
    s_{jl, t+1} = e^{i \Delta \omega_j} \cdot s_{jl, t} + y_{l,t}.
\end{align} 
The complex state $s$ of a neuron rotates around an angle $\Delta \omega_j$ per sampling time $\Delta t$ leading to an angular~velocity~\mbox{$\omega_j = \frac{\Delta \omega_j}{\Delta t}$}.
The sampling time $\Delta t$ and the chirp time $T_c$ define the number of samples $N$ for one chirp.
By iteratively applying Eq. (\ref{eq:discrete_neuron}) it can be shown, that the neuron state $s(T_c)$ at time $T_c$ with $N$ samples resembles the result of a DFT of the data vector \mbox{$y_l = (y_{0}, ..., y_{N-1})^T$} by setting $\Delta \omega_j = \frac{2 \pi jn}{N}$,
\begin{align}
\label{eq:fourier}
    s_{jl}(T_c) &=  e^{i N \Delta \omega_j} y_0 + e^{i (N-1) \Delta \omega_j} y_{l, 1} + ... +  e^{i \Delta \omega_j} y_{l, N-1} \nonumber \\
            &=  \sum_{n=0}^{N-1} e^{-i 2\pi jn/N} y_{l, n}.
\end{align}
Each neuron $s_{jl}$ is indexed by $l$ arising from the DFT along the antenna dimension. 
The phase shift $\phi_l$ represents an angle $\theta_l$.
Each neuron is also indexed by $j$ depending on the angular velocity $\omega_j$, which represents a range $r_j$.
Hence, each neuron is specific for a DoA angle $\theta_l$ and a range $r_j$.
Figure \ref{fig:architecture} visualizes the layout of the network architecture.
As depicted in (\ref{eq:fourier}), the state of the neuron at time $T_c$ represents the result of an FT.
In the next section, we work out neuron dynamics and spiking functions that emit spike trains before this state is reached while maintaining the accuracy of the FT. 

\subsection{Envelope estimation and gradient estimation}
\label{sec:gradient}

To estimate the angle and range of an object, we need to detect neurons that match the phase (\ref{eq:angle_detection}) and the frequency (\ref{eq:sol_de_neuron}) of an object. 
Matching frequencies (resonance) lead to small oscillations $\Delta \omega_{jk} = \omega_j - \omega_{k} \ll 1$ of the neuron state, and we apply Taylor-approximation on the magnitude of (\ref{eq:sol_de_neuron}) for a single target $k$ around $\Delta \omega_{jk}$,
\begin{align}
    |s_{jl}(t)| \simeq \beta_{kl} t^{1} + \mathcal{O}(t^2).
\end{align}
A matching phase of an object $k$ leads also to an increased amplitude $\beta_{kl}$ of the neuron state (\ref{eq:sol_de_neuron}), i.e. $\beta_{kl} \approx 1$.
Therefore, the first-order term of the magnitude of the neuron state stores the information of interest.
We assume that the remaining parts of the sum (non-resonant) affect only the offset and higher-order oscillations.

Ultimately, two criteria distinguish a neuron that matches the phase and frequency of an object and a non-matching neuron: (a) the magnitude $|s_jl(t)|$ follows a linearly increasing function $\beta_{kl} \cdot t$ over time and (b) the gradient of the linear function is large, $\beta_{kl} \gg 1$.

Applying this knowledge, we develop a neuron that estimates an envelope $\Lambda(t)$ of the magnitude of the neuron state by removing uninformative high-order oscillations.
We start by estimating the maximum of the neuron's magnitude $s_\text{max}(t)$,
\begin{align}
    &s_\text{max}(t) = \begin{cases}
        \|s(t)\|, &\text{ if } \|s(t)\| > s_\text{max}(t) \\
        s_\text{max}(t), &\text{ else.}
    \end{cases}
\end{align}

By estimating only the upper boundary, we neglect any decays or oscillations in the magnitude, which indicate non-resonating behavior.
Therefore, we also estimate the maximum width $w_\text{max}(t)$ between the neuron's magnitude and the upper boundary to extract the additional information on non-resonance.
The maximum of the width $w_\text{max}(t)$ is only updated when the width increases,
\begin{align}
    &w_\text{max}(t) = \begin{cases}
        s_\text{max}(t) - \|s(t)\|, &\text{ if } s_\text{max}(t) - \|s(t)\| > w_\text{max}(t) \\
        w_\text{max}(t), &\text{ else.}
    \end{cases}
\end{align}
The difference $\Lambda(t) = s_\text{max}(t) - w_\text{max}(t)$ follows the neuron's magnitude as a lower boundary and replicates an envelope estimate removing higher-order oscillations.

We estimate the gradient of $\Lambda(t)$ using exponential filtering,
\begin{align}
\label{eq:gradient}
   g(t) = (1-\alpha_g) g(t) + \alpha_g \Delta \Lambda(t)
\end{align}
$\Delta \Lambda$ being the update step for the envelope which depends on the updates in $s_\text{max}(t)$ and $w_\text{max}(t)$, \mbox{$\Delta \Lambda = \Delta s_\text{max} - \Delta w_\text{max}$}.

Figure \ref{fig:spiking_functions} shows the estimated gradient $g(t)$, $s_\text{max}(t)$, and $w_\text{max}(t)$. 
Table \ref{tab:neuron_model} summarizes all variables in the neuron model and indicates their dimensions, temporal behavior, and a short description.
For the first chirp, we set the initial value to $g(0)=0$. Otherwise, we use the last gradient estimate $g(T_c)$ as the initial value for the next chirp.
Instead of depending on the result of $s(T_c)$, as it replicates the result of an FT, we utilize information about the gradient estimate $g(t)$ to produce informative spikes.
This procedure allows us to already transmit information during the sampling process.

\subsection{Spiking Functions}
\label{sec:spiking_functions}

Neuromorphic applications rely on different spike encoding schemes, such as rate-coding and temporal-coding \cite{auge2021}.
In the following section, we describe three approaches that offer individual advantages and trade-offs.
The adaptive threshold spiking function builds on rate-coding and benefits from fewer computation steps, but transmits the most spikes.
The rate-coded LIF reduces the spike number and allows for early spike transmission, but increases the computational complexity. 
The temporal-coded LIF reduces the spike number even more, but does not benefit from early spike transmission.
Figure \ref{fig:spiking_functions} compares the dynamics and spiking behavior of all spiking functions for a single chirp. 
Table \ref{tab:spiking_functions} summarizes the variables and parameters of all the spiking functions.

\subsubsection*{Adaptive Threshold Spiking Function}

The first spiking function relies on an adaptive threshold proposed in \cite{auge2020selective}.
Once the neuron's state reaches a threshold $u_\text{th}$, the neuron spikes and increases this threshold by $\Delta u_\text{th}$. 
Therefore, the total number of spikes is proportional to the maximum value of the neuron within a time window, independent of the negative gradient within this window.
Since the magnitude of the neuron state $|s|$ oscillates in case of no resonance \mbox{$\omega_k \neq \omega_j$}, the decrease of the neuron state also contains information. 
Hence, we propose positive and negative spikes that depend on adaptive thresholds $u_\text{th}^{s}$ and $u_\text{th}^{w}$.
The neuron follows the spike generation rules,

\begin{align}
    \delta^+(t) = \begin{cases}
    1 & s_\text{max}(t) > u_\text{th}^s(t), \\
    0 & \text{else,}
    \end{cases}
\end{align}
and
\begin{align}
    \delta^-(t) = \begin{cases}
    -1 & w_\text{max}(t) > u_\text{th}^w(t),\\
    0 & \text{else,}
    \end{cases}
\end{align}
where $s_\text{max}$ and $w_\text{max}$ describe the estimated maxima in section \ref{sec:gradient}. 
The initial thresholds $u_\text{th}^{s/w}$ are updated according to

\begin{align}
    u_\text{th}^{s/w}(t) =  u_\text{th}^{s/w} + \Delta u_\text{th} \int_0^t \delta^\pm(t') dt' = u_\text{th}^\pm + \gamma N^{\pm}_\text{spikes} \, .
\end{align}

For the readout, the number of negative spikes is subtracted from the number of positive spikes $N^+_\text{spikes} - N^-_\text{spikes}$. 
This result gives an estimate of the envelope gradient $\Lambda(t)$ without calculating Eq. (\ref{eq:gradient}) directly. 
For consecutive chirps, the thresholds are reset to $\Gamma^\pm(0) = 0$.

\subsubsection*{Rate-coded LIF Spiking Function}

Alternatively, we can extend the neuron model using the dynamics of a LIF neuron model,
\begin{align}
    \tau \frac{du}{dt} = - u + g(t) + u_\text{rest} \, ,
\end{align}
to produce spikes with a spike rate $r(t) \sim g(t)$.
The LIF neuron spikes once the neuron membrane $u(t)$ reaches the threshold $u_\text{th}$ and resets its state by subtracting $u_\text{th}$. 
A constant gradient $g$ creates a spike rate
\begin{align}
    r(g) = \frac{1}{-\tau \ln(1- \frac{u_\text{th}}{g + u_\text{rest}})}.
\end{align}
In the case of no leak, the IF model creates a spike rate that is linearly proportional to $g$ within the time window $[0, T_c]$,
\begin{align}
    r(g) = \frac{g + u_\text{rest}}{u_\text{th}} \cdot \frac{T_c}{\tau} \,.
\end{align}
For consecutive chirps, the membrane is rest to $u(0) = 0$.

\subsubsection*{Time-coded LIF Spiking Function}

\begin{figure}
    \centering
    \includegraphics[width=\columnwidth]{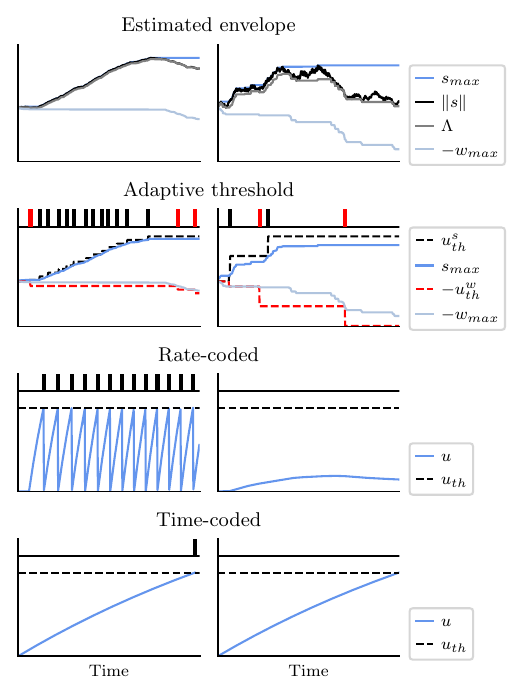}
    \caption{
    Comparison of neuron dynamics and spiking functions. The left column shows the neuron behavior when a target is present, whereas in the right column no target is present.
    The first row shows the the neuron's magnitude $\|s\|$, and its estimated maximum $s_\text{max}$ and the estimated maximum width $w_\text{max}$ between the magnitude and its maximum.
    The estimated envelope $\Lambda$ (grey) follows a lower boundary of the magnitude $\|s\|$.
    The second row shows the behaviour of the adaptive threshold spiking function. In the top row, positive spikes are indicated black and negative spikes red.
    A positive (negative) spike is generated when $s_\text{max}$ ($w_\text{max}$) reaches the threshold $u_\text{th}^{s}$ ($u_\text{th}^{w}$).
    The negative width $-w_\text{max}$ is shown for better visibility.
    The third row shows the behaviour of the rate-coded LIF spiking function. A present target results in a spike rate, whereas no target does not produce any spike.
    The fourth row shows the behaviour of the time-coded LIF spiking function. A present target results in a single spike, whereas no target does not produce any spike.
    }
    \label{fig:spiking_functions}
\end{figure}

Instead of transmitting the information about g(t) via the spike rate, we can use a more efficient encoding.
As in \cite{lopez2022time, lopez2023integrate}, a single spike encodes a constant value by charging a LIF model and using an adaptive refractory period.
Similar to the rate-coded approach, we calculate the spike time depending on the constant input $g$ as
\begin{align}
    t_s(g) = -\tau \ln(1-\frac{u_\text{th}}{g + u_\text{rest}}) \, .
\end{align}
Once the neuron spikes, we can read out the value $g$ by using a linear decoding scheme,
\begin{align}
    g \simeq T_c - t_s(g) \, .
\end{align}
Figure \ref{fig:spiking_functions} illustrates the charging of the neuron and the transmitted spike.
For consecutive chirps, the membrane is also reset to $u(0) = 0$.
Table \ref{tab:spiking_functions} summarizes the variables and parameters of the time-coded spiking function and compares them with the alternative spiking functions.

\begin{table}
\caption{Neuron states and parameters for neuron model excluding spiking functions}
\label{tab:neuron_model}
\begin{tabular}{r l l l}
    \multicolumn{1}{c}{Variable} & \multicolumn{1}{c}{Dim.} & \multicolumn{1}{c}{Time} & \multicolumn{1}{c}{Description} \\
    \hline
     $\textbf{x}(t)$        & $r$ & dyn.    & radar data from $N_\text{vx}$ antennas\\
     $y(t)$                 & $1$ & dyn.    & phase-shifted radar data $y(t) = \vec{w}\vec{x}(t)$ \\
     $\textbf{s}(t)$        & $2$ & dyn.    & neuron state of RF neuron\\
     $\|s(t)\|$             & $1$ & dyn.    & magnitude of neuron state\\
     $s_\text{max}(t)$               & $1$ & dyn.    & estimated maximum of magnitude\\
     $w_\text{max}(t)$               & $1$ & dyn.    & \makecell{estimated maximum of sum of width between\\
                    magnitude and its maximum}\\
     \hline
     $\textbf{w}$         & $r$ & static  & weight vector of complex weight matrix $W$ \\
     $\omega$             & $1$ & static  & eigenfrequency  \\
\end{tabular}
\end{table}

\begin{table}
\caption{Neuron states and parameters for spiking functions}
\label{tab:spiking_functions}
\begin{tabular}{lll| lll| lll}
     \multicolumn{3}{c|}{Adaptive threshold} & \multicolumn{3}{c|}{Rate-coded LIF} & \multicolumn{3}{c}{Time-coded LIF} \\
     \hline
     Var. & Dim. & Time & Var. & Dim. & Time & Var. & Dim. & Time \\
     \hline
     -               & -    & -     & $g(t)$    & $1$ & dyn. & $g(t)$ & $1$ & dyn.\\
     $\Gamma^\pm(t)$ & $2$  & dyn.  & $u(t)$    & $1$ & dyn. & $u(t)$ & $1$ & dyn.\\
     \hline
     -               & -    & -     & $\alpha_g$            & $1$ & static& $\alpha_g$            & $1$ & static\\
     $\gamma$        & $1$  & static& $u_\text{th}$  & $1$   & static& $u_\text{th}$  & $1$ & static\\
     -               & -    & -     & $u_\text{rest}$       & $1$ & static& $u_\text{rest}$       & $1$ & static\\
     -               & -    & -     & $\tau$                & $1$ & static& $\tau$                & $1$ & static\\
\end{tabular}
\end{table}

\section{Evaluation}
\label{sec:evaluation}

The evaluation of temporal neuron models is time- and energy-consuming, as the processor needs to calculate each time step of each neuron.
Therefore, we implemented our neuron model on GPU to process all neurons of the range-angle map in parallel. 
Each neuron runs independently on a single CUDA core, removing waiting or communication time between different cores. 
This parallelization drastically improved the computation time by a factor $\sim 1000$ and allowed us to evaluate our network model on large amounts of data. 

\subsection{Dataset Simulation}

For the evaluation, we generated raw radar data from a radar simulator by Infineon Technologies AG. 
Table~\ref{tab:sim_radar_params} summarizes the parameters of the simulated radar sensor. 
The simulation includes phase noise, thermal noise, and phase-locked loops.
The simulator creates targets as single points with a given position, velocity, and radar cross-section (RCS). 
With these parameters, we created a labeled dataset consisting of the raw radar data and a list of point targets, including their attributes.
Using two different seeds, we simulated five scenarios, summarized and described in Table~\ref{tab:sim_datasets}. 
The dataset with seed 0 will be used as a training dataset to optimize the parameters, whereas seed 1 is the evaluation dataset.

\begin{table}
    \centering
    \caption{Parameters of the simulated radar sensor}
    \begin{tabular}{ll}
         \multicolumn{1}{c}{Parameter name} & \multicolumn{1}{c}{Parameter value} \\
         \hline
         initial frequency $f_0$ &  $76~\text{GHz}$\\
         bandwidth $B$ &  $507.6~\text{MHz}$\\
         number of samples $N_\text{samples}$ & $512$\\
         number of chirps $N_\text{chirps}$ & $32$\\
         number of receiving antennas $N_\text{rx}$ & $32$\\
         number of transmitting antennas $N_\text{tx}$ & $1$\\
         duration of chirp $t_\text{chirp}$ & $20.52~\mu s$ \\
         duration of wait time $t_\text{wait} $& $5.96~\mu s$\\
         chirp frequency $f_\text{chirp}$& $37.76~\text{kHz}$\\
    \end{tabular}
    \label{tab:sim_radar_params}
\end{table}

\begin{table}
    \centering
    \caption{List of simulated data sets}
    \begin{tabular}{ll}
         \multicolumn{1}{c}{Dataset name} & \multicolumn{1}{c}{Description} \\
         \hline
         \textit{close targets 2010}          & \makecell{two targets close by with \\
         $\sigma_0 = 10$ and $\sigma_1 = 20$}\\
         \textit{close targets 0010}          & \makecell{two targets close by with \\
         $\sigma_0 = 10$ and $\sigma_1 = 00$}\\
         \textit{5 mixed objects} & \makecell{five targets randomly distributed with \\
         $\vec{\sigma} = [0, 5, 10, 15, 20]$}\\
         \textit{5 persons}   & \makecell{five targets randomly distributed with \\
         $\vec{\sigma} = [0, 0, 5, 5, 10]$}\\
         \textit{8 targets}                    & up to 8 targets randomly distributed
    \end{tabular}
    \label{tab:sim_datasets}
\end{table}

\subsection{Metric}

Since our model resembles the concept of frequency analysis, we compare our results with the FT. 
To evaluate the performance of our model, we measure the signal-to-noise ratio (SNR) and the accuracy of the target detection by applying the CA-CFAR algorithm \cite{barkat1987} to the neurons' output to create classification maps. 
These maps indicate whether a target is present at a given frequency bin for a distance or an angle.
The two approaches, SNN and FT, are compared by their SNR, F-score, precision, and recall values.

The sparsity of the SNN approaches leads to a majority of zeros in the range-angle map. 
Hence, we slightly adjust the calculation of the SNR by dividing the signal, i.e., the data point $d_\text{target}$ at the range and angle index of a target, by the sum of data points in the whole map to avoid divisions by zero,
\begin{align}
    \text{SNR}_\text{target} = \frac{d_\text{target}}{\sum_{ij} d_{ij}}.
\end{align}
This adjustment leads to $\text{SNR} = 1$ for no noise.
For simplicity, we fix the number of neighboring cells and use no guard cells for all tests. 
Hence, we parameterize the CA-CFAR algorithm by only two parameters: an offset $o$ that adds to the threshold for the cell under test (CUT) derived from the window and a weight factor $\alpha$ that scales the resulting threshold.
The threshold for CUT $c_{ij}$ is given by
\begin{align}
    \Theta_\text{CUT}(c_{ij}) = \alpha \cdot \left(\frac{1}{K} \sum_{kl}^{K} c_{kl} + o \right) \, ,
\end{align}
with ${K}$ as a set of neighboring cells. 
We fixed $K= 3 \times 5$ for all experiments in the range and angle dimensions, respectively.
Typically, the SNN output contains mainly zeros. Therefore, the offset $o$ is needed to generate useful results of the CA-CFAR algorithm.

\subsection{Parameter optimization}

We distinguish two parts in our neuron model: the gradient estimation and the spiking function. 
To tune the parameters of the whole neuron model, we optimized the two parts consecutively. 
First, we optimized the parameters of the gradient estimation and CA-CFAR using a grid search on a training data set by maximizing the F-score. 
Second, we set the gradient estimation parameters and optimized the spiking functions' parameters with the same approach.

\subsection{Model evaluation for a single chirp}

The three spiking functions were evaluated and compared to the FT results. 
In addition, we analyzed the results of the gradient estimation by directly using the value of the estimated gradient. 
Table~\ref{tab:optim_params_1chirp} summarizes the optimized parameters of each model, and Table~\ref{tab:results_1chirp} shows the F-score, precision, and recall for each spiking model and the comparable FT approach over all datasets.
These results show that the rate- and time-coded spiking functions reach higher F-scores than the gradient or adaptive threshold model.
Because the results of the gradient model and FT are similar, we can conclude that the advantage of the two spiking models mainly lies in the non-linear mapping of the spiking function itself. 
The adaptive threshold spiking function, which utilizes a linear mapping from values to spikes, shows only a slight loss in the F-score, affirming the previous assumption.
An intrinsic cut-off can explain the increased SNR of all neuron models, as only estimated gradients larger than $0$ are transmitted.
Even though our model eliminates sensor data storage by continuous processing and transmits only spikes, it performs similarly in classification accuracy as the FT.
The intrinsic cut-off and the usage of binary spikes drastically reduce the data bandwidth, as the number of generated spikes shows.
The entire network consists of $256 \times 32 = 8192$ neurons.
A network utilizing the adaptive threshold spiking function transmits, on average, $27128$ spikes, which is more than the other spiking functions, as it sends positive and negative spikes.
The rate-coded spiking function transmits $910$ spikes per network, leading already to less than one spike per neuron on average.
By utilizing the time-coded spiking function, the entire network transmits, on average, only $53$ spikes.
In contrast, the classic FT approach transmits $256 \times 32 = 8192$ floating values, i.e., $262\, 144$~bits per chirp, assuming float-$32$.
The spiking neural resonator network transmits only 0.02 $\%$ of the data transmitted by the classic float-32 FT approach.

\begin{table}
\caption{Optimized parameters of the spiking and non-spiking gradient models for the single chirp use-case}\def\arraystretch{1.2}
\begin{tabular}{l l}
     \multicolumn{1}{c}{Model name} & \multicolumn{1}{c}{Optimized parameters} \\
     \hline
     gradient model & $\alpha_x = 0.6, \alpha_g = 0.001$\\
     adaptive threshold & $\gamma= 0.1$ \\
     rate-coded LIF     & $u_\text{threshold}=0.35, u_\text{rest}=0, \tau=100$ \\
     time-coded LIF     & $u_\text{threshold}=231, u_\text{rest}=250, \tau=200$ \\
\end{tabular}
\label{tab:optim_params_1chirp}
\end{table}

\begin{table}
\caption{Evaluation results on simulated data after a single chirp}
\begin{tabular}{l l l l l l l}
     \multicolumn{1}{c}{Model name} & F-score & Prec. & Recall & SNR & avg. \# spikes \\%$\frac{\text{\# spikes}}{\text{neuron}}$\\
     \hline
     gradient model     & \textbf{0.61} & \textbf{0.66} & 0.58 & 0.012 &  \\ 
     adaptive threshold & 0.60 & 0.65 & 0.58 & 0.010 & $\sim 27128$ \\  
     rate-coded LIF     & \textbf{0.61} & 0.59 & 0.64 & 0.104 & $\sim 910$ \\  
     time-coded LIF     & \textbf{0.61} & 0.57 & \textbf{0.66} & \textbf{0.105} & $\sim \textbf{53}$ \\  
     FT                 & 0.59 & 0.64 & 0.56 & 0.006 & \\ 
\end{tabular}
\label{tab:results_1chirp}
\end{table}

\subsection{Model evaluation of early detections for a single chirp}

\begin{figure*}[!h]
    \centering
    \includegraphics[width=\textwidth]{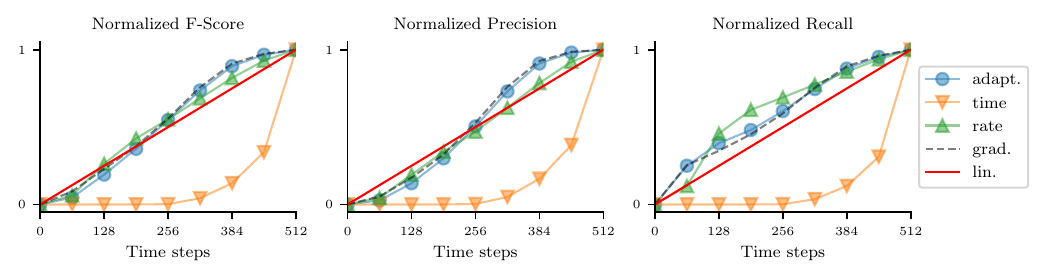}
    \caption{
    Early detection evaluation for adaptive threshold (adapt.), time-coded (time), rate-coded (rate), and the gradient (grad) model. 
    After every $64$ data sample, we determine the neural resonator network's F-score, precision, and recall. 
    The maximum value of each model normalizes the values. 
    For comparison, a linear reference is shown in red. 
    The F-Score of the adaptive threshold, rate-coded, and gradient model increases almost linearly, as the resolution of the Fourier transform also increases linearly with the number of samples. 
    The time-coded model shows a different behavior, as most informative spikes happen at the end of a chirp.
    }
    \label{fig:early_detection}
\end{figure*}

Because our neuron model continuously processes sensor data and transmits information by sending spikes, objects are already detected during data sampling.
In contrast, the FT relies on sampling and storing the data before performing any calculation.
Therefore, we also studied the early detection accuracy of our neuron model by evaluating the spike output at every $64$th time step to analyze whether early predictions of close-by objects are possible.
We performed the evaluation on the \textit{close~targets~2010} dataset.
The recall results depicted in Figure \ref{fig:early_detection} show that for the adaptive threshold and rate-coded spiking functions already after half of the available samples $N_\text{samples}/2$ more than $75$ \% of the detectable objects are recognized. 
The time-coded spiking function, however, needs due to its internal dynamics more time to produce a spike. Since most of the peaks in the frequency spectrum are close to zero, the spikes typically happen at the end of a chirp.

\subsection{Model evaluation for multiple consecutive chirps}

Due to a high chirp rate ($\sim$ kHz) of FMCW radar sensor \cite{fmcw_radar_designs} changes in the scenes between consecutive chirps are hardly detected in range-angle maps.
Consequently, it is natural to exploit the continuous computation capabilities of the neurons to process consecutive chirps without resetting the entire network.
Therefore, we also evaluate the classification accuracy after processing multiple chirps of the same scene. 
We follow the optimization strategy described before and summarize the new parameters for processing $8$ chirps in Table \ref{tab:optim_params_8chirps}. 
It is important to emphasize that the optimal parameters depend not directly on the number of processed chirps but on the state of convergence to the gradient estimation. 
After a specific number of chirps $\tilde{N}_\text{chirps}$, the estimated gradient converges.
By optimizing the parameters with the assumption of a converged gradient estimation, results are optimal for all chirps $n_\text{chirps} > \tilde{N}_\text{chirps}$. 
Earlier processed chirps will result in reduced accuracy.

The results summarized in Table \ref{tab:results_8chirps} show an increased performance compared to the single chirp evaluation.
These results are also compared to an evaluation of averaging the output over the same $8$ consecutive chirps.
In this case, the same parameters of the single chirp evaluation are taken (Table \ref{tab:optim_params_1chirp}).
Table \ref{tab:results_8chirpsavg} summarizes the results for F-Scores, precision, recall, SNR, and spike numbers, including the results of averaging the FT calculations.
Since every single chirp produces similar spike numbers, the average number is higher than the continuous processing of chirps in Table \ref{tab:results_8chirps}. 
The detection accuracy of the neuron models is comparable and situated between the continuous processing and averaging approach, leading to the conclusion that storing outputs of multiple chirps is unnecessary. 
In general, the FT approach performs slightly worse than all neuron models.

\begin{table}
\caption{Optimized parameters of the spiking and non-spiking gradient models for the consecutive chirps use-case}
\begin{tabular}{l l}
     \multicolumn{1}{c}{Model name} & \multicolumn{1}{c}{Optimized parameters} \\
     \hline
     gradient model & $\alpha_x = 0.6, \alpha_g = 0.001$\\
     adaptive threshold & $\gamma= 0.1$ \\
     rate-coded LIF     & $u_\text{threshold}=1.5, u_\text{rest}=0, \tau=100$ \\
     time-coded LIF     & $u_\text{threshold}=232, u_\text{rest}=250, \tau=200$ \\
\end{tabular}
\label{tab:optim_params_8chirps}
\end{table}

\begin{table}
\caption{Evaluation results on simulated data after eight consecutive chirps}
\begin{tabular}{l l l l l l}
     \multicolumn{1}{c}{Model name} & F-score & Prec. & Recall & SNR & avg. \# spikes \\ 
     \hline
     gradient model     & \textbf{0.72} & \textbf{0.73} & 0.71 & 0.013 & \\ 
     adaptive threshold  & 0.60 & 0.65 & 0.58 & 0.010 & $\sim 2579$ \\  
     rate-coded LIF     & 0.69 & 0.67 & \textbf{0.72} & \textbf{0.117} & $\sim 126$\\ 
     time-coded LIF     & 0.68 & 0.66 & 0.71 & 0.103 & $\sim \textbf{5}$\\  
\end{tabular}
\label{tab:results_8chirps}
\end{table}

\begin{table}
\caption{Evaluation results on simulated data averaging eight consecutive chirps.}
\begin{tabular}{l l l l l l}
     \multicolumn{1}{c}{Model name} & F-score & Prec. & Recall & SNR & avg. \# spikes \\ 
     \hline
     gradient model     & \textbf{0.72} & 0.71 & \textbf{0.72} & 0.012 & \\ 
     adaptive threshold  & 0.64 & \textbf{0.75} & 0.56 & 0.010 & $\sim 20632$\\ 
     rate-coded LIF     & 0.63 & 0.59 & 0.67 & \textbf{0.105} & $\sim 372$\\ 
     time-coded LIF     & 0.63 & 0.59 & 0.68 & \textbf{0.105} & $\sim \textbf{43}$\\ 
     FT                 & 0.60 & 0.63 & 0.59 & 0.006 & \\  
\end{tabular}
\label{tab:results_8chirpsavg}
\end{table}

\subsection{Visual results on real data}
\label{sec:real_data}

\begin{figure*}[!h]
    \includegraphics[width=\textwidth]{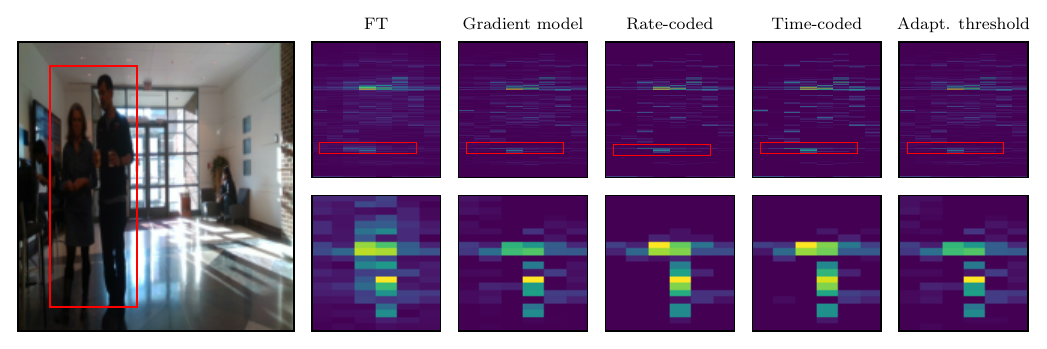}
    \caption{
    Comparison of range-angle maps from FMCW radar sensor data of the RADICAL dataset for one scene.
    (Left) RBG image of the RADICAL dataset of the corresponding scene.
    (Right, Top) Full-resolution range-angle maps showing the full spectrum. 
    (Right, Bottom) Zoomed-in extract of the two people detected by the radar sensor.
    The noise level of the range-angle map of the Fourier Transform is higher than the non-spiking and spiking range-angle map of neural resonator networks. 
    The gradient and adaptive threshold spiking function maps are similar, as are the rate-coded and time-coded maps.
    }
    \label{fig:radical}
\end{figure*}

To show the applicability of our model on real-world radar sensor data, we applied it to publicly available raw radar data. 
The availability of raw radar data including a multi-antenna layout is limited.
The RADICAL dataset consists of a sample data file of indoor recordings of walking persons. 
Figure \ref{fig:radical} shows a scene containing $4$ persons, two walking close to the sensor.
We compare the range-angle maps of the FT calculation, the gradient model, and the three different spiking functions.
The comparison shows a similar frequency spectrum for all approaches (upper row in Fig. \ref{fig:radical}). 
However, our models show a significantly reduced noise level, which becomes evident in the zoomed-in extracts of the range-angle maps (lower row in Fig. \ref{fig:radical}).

\section{Conclusion}

This work extends the resonate-and-fire model to optimize it for radar processing applications.
We focused on reducing latency and data bandwidth by allowing spikes during the sampling of radar data, eliminating the need to store any sensor data. 
Our continuous approach, therefore, stands out in contrast to the classic Fourier transformation, where all data points are sampled, stored, and processed afterward.
In general, utilizing only a subset of all samples within a chirp reduces the resolution of the frequency analysis.
However, the neuron model spikes during sampling time; we have shown that the object detection accuracy is comparable to or even better than the FT approach.
This is achieved by first estimating the first-order part (linear in time) of the resonator dynamics, which filters out noise from non-resonating objects, and second, by emitting spikes proportional to the gradient of the linear part, which is comparable to a threshold function since only positive gradient above a specific value can produce spikes.
The three different spiking functions compared in this work show similar target detection accuracy but have several trade-offs one must consider.
The adaptive threshold spiking function is the most straightforward of these three but transmits the most spikes, including negative ones.
The rate-coded spiking function has a similar complexity as the time-coded function but uses more spikes to convey information and is therefore suitable for early detection. 
The time-coded spiking function utilizes the least number of spikes but lacks the feature of early detection.
To conclude, we have demonstrated that a network based on our spiking neural resonators can achieve the same performance as the FT while reducing the latency and data bandwidth.
The data bandwidth can be reduced to 0.02 $\%$ of the data bandwidth for a float-32 FT.

\section{Future work}

The current state of the network does not implement any lateral connections; therefore, each neuron processes information independently, allowing complete parallel computations and optimizations.
Introducing laterally inhibiting connections between neurons can theoretically improve the detection accuracy by suppressing neighboring neurons excited by the same target.

So far, we evaluated a static data set, i.e. the targets were not moving, and we did not consider temporal changes. 
The continuous setup of the neuron model does not reset the internal variable of the gradient estimation and uses the information of previous chirps. 
Therefore, we plan to evaluate the performance and the reaction time of the network in a highly dynamic and realistic environment. 

The neuron model is continuous in time and can theoretically process analog data directly from the radar sensor. 
Analog processors that implement this model can be highly efficient in energy consumption and should be explored in the future. 
First implementations of analog resonate-and-fire models have been studied in \cite{lehmann2023}.
Directly processing analog signals and transmitting only sparse spikes for further post-processing paves the way for efficient neuromorphic radar sensors.

\section*{Acknowledgments}
This research has been funded by the Federal Ministry of Education and Research of Germany in the framework of the KI-ASIC project (16ES0995).

\bibliographystyle{plain}
\bibliography{main.bib}

\end{document}